\documentclass[journal]{IEEEtai}

\usepackage[colorlinks,urlcolor=blue,linkcolor=blue,citecolor=blue]{hyperref}

\usepackage{color,array}

\usepackage{graphicx}

\usepackage{graphicx}
\usepackage{subfig}
\usepackage{multicol}
\usepackage{booktabs, multirow} 
\usepackage{soul}
\let\labelindent\relax
\usepackage{enumitem}
\usepackage{todonotes}
\usepackage{hyperref}

\begin{document}

\title{Exploiting Representation Bias for Data Distillation in  Abstractive Text Summarization}

\author{
    Yash Kumar Atri$^{1}$, Vikram Goyal$^{1}$, Tanmoy Chakraborty$^{2}$  \\
    $^{1}$IIIT-Delhi, New Delhi, India; $^{2}$IIT Delhi, New Delhi, India \\
    {\tt \{yashk,vikram\}@iiitd.ac.in; tanchak@iitd.ac.in}
}


\maketitle

\begin{abstract}
Abstractive text summarization  is surging with the number of training samples to cater to the needs of the deep learning models. These models tend to exploit the training data representations to attain superior performance by improving the quantitative element of the resultant summary. However, increasing the size of the training set may not always be the ideal solution to maximize the performance, and therefore, a need to revisit the quality of training samples and the learning protocol of deep learning models is a must. In this paper, we aim to discretize the vector space of the abstractive text summarization models to understand the characteristics learned between the input embedding space and the models' encoder space. We show that deep models fail to capture the diversity of the input space. Further, the distribution of data points on the encoder space indicates that an unchecked increase in the training samples does not add value; rather, a tear-down of data samples is highly needed to make the models focus on variability and faithfulness. We employ clustering techniques to learn the diversity of a model's sample space and how data points are mapped from the embedding space to the encoder space and vice versa. Further, we devise a metric to filter out redundant data points to make the model more robust and less data hungry. We benchmark our proposed method using quantitative metrics, such as Rouge, and qualitative metrics, such as BERTScore, FEQA and Pyramid score. We also  quantify the reasons that inhibit the models from learning the diversity from the varied input samples. 
\end{abstract}

\begin{IEEEImpStatement}
In the realm of abstractive text summarization, the surge in training data for deep learning models promises superior quantitative summary performance. However, blindly increasing the dataset size does not consistently yield optimal results. This necessitates a reevaluation of training sample quality and learning strategies. Our study dissects the vector space of abstractive text summarization models, revealing that deep models struggle to capture input space diversity. Furthermore, excessive training data fails to provide substantial value. We employ clustering techniques to understand sample space diversity and devise a metric to filter redundant data, enhancing model robustness while reducing data dependency. Benchmarking against quantitative and qualitative metrics corroborates our approach. This research not only identifies critical model limitations but also offers a pragmatic solution, ushering in an era of more effective, data-efficient summarization techniques.
\end{IEEEImpStatement}


\section{Introduction}

Over the past years, summarization systems have evolved from traditional graph-based approaches \cite{erkan2004lexpagerank, textrank2004, 10.1145/3130348.3130369,favre2008tac} to machine learning and deep learning based systems \cite{nallapati-etal-2016-abstractive,see2017get}. The recent advancements in attention-based architectures \cite{DBLP:journals/corr/VaswaniSPUJGKP17} have further pushed the performance at the cost of large labelled datasets. Despite the huge success, these systems often fail to maintain consistency in performance across multiple datasets \cite{dey-etal-2020-corpora}. This inconsistency is observed primarily in the qualitative parameters like generalizability, faithfulness and coherence. The factors like the diversity of users, community theme, and moderation can be accounted for quality degradation arising from scrapping crowdsourcing platforms. Intuitively, the diversity and variability should help the model generalize better; however, the skewed representation of the training samples abstains the systems from learning variable representations (Table \ref{table:sample}).

Deep learning based abstractive text summarization models exploit the schematic representations in the training data to capture the relationship between the source and the target. However, for a given dataset, a large portion of the training sample follows a similar structure in terms of style and feature representation making the systems learn it as an essential representation while ignoring the other variable samples as noise. This imbalance of features in the corpus leads to representation bias where the majority features get importance while minority features are considered outliers. This skewed representation of samples in the training data incapacitates the model from understanding the varied representation across the input space, making it learn over a saturated sample leading to degraded performance. The models trained on these skewed data generate repetitive, templatic and unfaithful \cite{Holtzman2020The,li-etal-2019-conclusion,fan-etal-2018-hierarchical} summaries. 

Representation bias is studied widely for binary and multi-class classification tasks but is often not given importance for language generation tasks like summarization. This representation bias in the training data causes the invariability in the vector space generated by the system, leading to degraded performance. In parallel, the system's training dynamics also contribute equally to relegating the quality of the output. Recently, several studies investigated biases in sequence-to-sequence (seq2seq) based networks \cite{zhang-etal-2020-minimize,mueller-etal-2022-coloring,kharitonov2021what}. \cite{zhang-etal-2020-minimize} minimized the exposure bias by limiting the decoding length; \cite{mueller-etal-2022-coloring} used finetuning language models; and \cite{kharitonov2021what} showed how seq2seq systems are biased towards arithmetic, and transformer based systems are biased for reasoning. However, the focus on bias identification has been limited to either study on dataset quality or identifying biases reflecting unfaithfulness, hallucinations or syntactical performance. Although the system's faithfulness and accuracy play vital roles, the diversity and generalizability of a system make the systems robust and practical to use. 

In this work, we characterize how the representation bias in the dataset influences the performance of the deep learning based abstractive text summarization systems. We examine this by studying the systems' overall training dynamics at the embedding and the encoder levels over various datasets. Next, we study the reasons which influence the diversity and generalizability of the systems. We argue that a system can be more faithful and robust without affecting the quantitative performance of the overall system. We also make an attempt to quantify the reasons that make the deep learning based models unfaithful, hallucinative, and repetitive.
 
 \begin{table}[t]
    \centering
    \small
        \begin{tabular}{p{8.5cm}}
        \hline
          \textbf{Source}: (...) manhattan building is the former site of an 18th-century tavern where \textcolor{black}{george washington is believed to have enjoyed a celebratory drink} during the american revolution . (...) george washington and governor (george) clinton stopped at the bull (...) fate of the site depends on the current owner , alex chu , who is \textcolor{black}{demolishing the site to make way for a new hotel} \\ \hline
          \textbf{Baseline}: \textcolor{red}{george washington} \textcolor{blue}{is believed believed have  enjoyed a celebratory} , \textcolor{red}{george washington}  and \textcolor{red}{washington} stopped is \textcolor{blue}{demolishing the site to make way for a new hotel .} \\ \hline
          \textbf{Exploited}: where \textcolor{blue}{george washington is believed to have enjoyed a celebratory drink during the american revolution} , the site is a " incredible opportunity to get on archaeological record . if so , this would be the oldest surviving structure in manhattan by far . \\ \hline
          \textbf{Gold}: george washington is believed to have stopped for a celebratory drink at the bull ' s head tavern . now he ' s hoping city officials will preserve the site , saying , " what an incredible opportunity that the city suddenly has for this thing to re-emerge .  \\ \hline
        \end{tabular}
    \caption{Source, generated, and  gold instances from the pointer-generator model \cite{see2017get}. Exploited summary are generated from the model trained on the sampled data. The red color highlights the repetition, while the blue color highlights the copying.}
    \label{table:sample}
\end{table}

 We quantify the representation bias in the dataset by moulding the training dynamics of the systems on three widely popular datasets -- CNN/Dailymail \cite{hermann2015teaching}, Multinews \cite{Multinews2019}, and CQASumm \cite{chowdhury2019cqasumm}, and six deep learning based models -- Pointer Generator \cite{see2017get}, PG-MMR \cite{lebanoff-song-liu:2018}, HiMAP \cite{Multinews2019}, Transformer \cite{DBLP:journals/corr/VaswaniSPUJGKP17}, CopyTransformer \cite{gehrmann2018bottom}, and Longformer \cite{Beltagy2020Longformer}.

Our major contributions are four-fold:
\begin{itemize}
    \item We explore the representation bias in the dataset during training by studying the training dynamics of six widely-popular deep learning based systems.
    \item We devise a metric to filter out redundant data samples that directly affect the performance of the system.
    \item We assess the quality of the filtered data points by benchmarking the performance of the system quantitatively as well as qualitatively on various metrics.
    \item We reason about the biases reflected by the dataset during the training and inference. 
\end{itemize}

\section{Related Works}
The task of summarization has improved over time with the advancements in neural network architectures. Sequence-to-sequence \cite{sutskever2014sequence}, attention-based networks \cite{nallapati-etal-2016-abstractive, ATRI2021107152, 10.1145/3580305.3599830, 10261260, atri-etal-2023-promoting}, and Transformers \cite{DBLP:journals/corr/VaswaniSPUJGKP17} are the few which led the regime of summarization to a much higher level. However, many of these studies in summarization focused on gaining performance on the datasets rather than quantifying the biases that may lead to the deterioration of the qualitative performance. Previous studies in characterizing biases in summarization include \cite{hirao2004corpus} using numerous intrinsic metrics involving length limits on summaries, precision, coverage and extrinsic metrics like question-answering (QA) based evaluation to assess faithfulness. \cite{benikova2016bridging} used inter-annotator agreement scores to model the quality of the annotated dataset. \cite{https://doi.org/10.48550/arxiv.2106.10084} studied the subjective bias in summarization by discussing the impact of having one and many target summaries for a single-source document. Recently, \cite{jung2019earlier} studied the quality using locality, diversity and informativeness. \cite{dey-etal-2020-corpora} studied both the datasets and systems over various intrinsic and extrinsic metrics. \cite{peyrard2019simple} defined bias in the form of relevance, importance and redundancy. 

Further studies have also been proposed to address biases on the system's side. \cite{cao-etal-2020-factual} and \cite{chen-etal-2021-improving} used post-correction techniques to make the generated summaries more faithful.  \cite{cao-etal-2020-factual} utilized synthetic data to learn the factual dependencies while \cite{chen-etal-2021-improving} learned a discriminative model to generate multiple system summaries and ranked them as per correlation with the source document. \cite{9739885} utilized granular information and fact-based attention to map the factual information. Over the dataset quality, \cite{https://doi.org/10.48550/arxiv.2204.10290} proposed revisions of reference summaries to remove substandard samples that affect the quality of the models. \cite{https://doi.org/10.48550/arxiv.2205.12854} mapped the error-centric labels and argued that no single system could correct all the known biases and errors.

However, most of the studies in identifying bias in summarization are restricted to studies pertaining to subjective bias, dataset mislabeling or factual and hallucinative biases. Existing literature uses various intrinsic and extrinsic metrics to capture these biases and propose new systems to mitigate them. The proposed systems use penalties and additional neural network modules to help the model mitigate these biases. However, the core reason that amplifies these biases remain unidentified. In this paper, we aim to study the factors that affect the systems' performance and generalizability. Further, we show that the performance of the system can be made more faithful and robust without affecting the syntactical performance. We also make an attempt to quantify the reasons that make the deep learning based summarization models unfaithful, hallucinative, and repetitive.

\begin{table*}[!t]\centering
\small

\scalebox{1.0}{
\begin{tabular}{lrrrrrrrrr}\toprule
Dataset &Train/val/test &Type & \#a-src &\#a-sum &Inter-sim &Pyr &Compr &Abstr \\\cmidrule{1-9}
CNN/Dailymail &286817/13368/11487 &News &810 &53 &4.68 &0.27 &15.28 &71.98 \\\cmidrule{1-9}
Multinews &44900/5622/562 &News &2103 &260 &2.37 &0.4 &8.08 &80.45 \\\cmidrule{1-9}
CQASumm &80000/10000/10000 &QA &784 &65 &8.25 &0.05 &12.06 &88.79 \\\midrule
\bottomrule
\end{tabular}}
\caption{Comparison of datasets, namely CNN/Dailymail, Multinews and CQASumm, over statistical metrics -- number of documents in train, validation and test sets, type of the dataset, average number of words (\#a-src) in the source document and the summary (\#a-sum), inter-document similarity score (Inter-sim), pyramid score (Pyr), compression ratio (Compr) and the 3-gram abstractness (Abstr).}\label{tab:datastats}
\end{table*}

\section{Experimental Setup}
Here we discuss the benchmarked datasets and the deep learning based abstract text summarization systems. Table \ref{tab:datastats} shows the average number of words in the source documents, average number of words in the summary, inter-similarity score representing the thematic overlap within the document, Pyramid score \cite{pyramid-m} and 3-gram abstractness of the datasets. Since the deviation in the average scores is less than 5\%, we combine the average scores of LSTM and Transformer-based models. We refer to Pointer Generator, PG-MMR,  and Himap as {\em LSTM-based models} and Transformer, CopyTransformer, and Longformer as {\em Transformer-based models}.

\subsection{Abstractive Text Summarization Datasets}
We benchmark our study on three  widely popular summarization datasets -- CNN/Dailymail \cite{hermann2015teaching}, Multinews \cite{Multinews2019}, and CQASumm \cite{chowdhury2019cqasumm}.
\begin{itemize}
    \item \textbf{CNN/Dailymail} \cite{hermann2015teaching}contains source documents scrapped from CNN/ Dailymail, and the target summary is annotated as the highlights of the news article written by professional editors. On average, the source documents consist of $781$ tokens, while the reference summaries consist of $56$ tokens. 
    \item \textbf{Multinews} \cite{Multinews2019} constitutes source documents as a news item, and the target summary is annotated by crowdsourcing platforms. On average, the source documents consist of $2k$ tokens, while the reference summaries consist of $260$ tokens. 
    \item \textbf{CQASumm} \cite{chowdhury2019cqasumm} is framed from the Yahoo! L6 dataset comprising question answers. The accepted answer is treated as a reference summary, while all other answers act as a source document. The source documents contain $784$ tokens, while reference summaries contain $65$ tokens. 
\end{itemize}


\subsection{Abstractive Text Summarization Systems}
\begin{itemize}
    \item \textbf{Pointer Generator} (PG) \cite{see2017get} is a bidirectional LSTM based seq2seq network in which the pointing mechanism allows the network to copy from the source document directly. 
    \item \textbf{PG-MMR} \cite{lebanoff-song-liu:2018} introduces maximal marginal relevance in the PG network to allow better coverage of information and reduce redundancy. 
    \item \textbf{HiMAP} \cite{Multinews2019} expands the PG network and uses MMR; the MMR scores are integrated into token-level attention weights during summary generation. 
    \item \textbf{Transformer} \cite{DBLP:journals/corr/VaswaniSPUJGKP17} uses self-attention to compute context relative to other tokens. The parallelization and the attention make the Transformer attend to all tokens and provide fast inference. 
    \item \textbf{CopyTransformer} \cite{gehrmann2018bottom} uses the standard Transformer \cite{DBLP:journals/corr/VaswaniSPUJGKP17} architecture. However, one of the random attention heads acts as copy pointer. 
    \item \textbf{Longformer} \cite{Beltagy2020Longformer} extends the vanilla Transformer \cite{DBLP:journals/corr/VaswaniSPUJGKP17} by incorporating windowed attention and global attention.
\end{itemize}

\subsection{Evaluation Metrics}
\begin{itemize}
    \item \textbf{Rouge} \cite{lin-2004-rouge} measures the $n$-gram overlap between the reference summary and the generated summary. We use Rouge-1, Rouge-2, Rouge-L and Rouge-L F1 for evaluating the performance of the summaries.
    \item \textbf{BERTScore} \cite{bert-score} uses the pre-trained BERT \cite{devlin-etal-2019-bert} and compares the cosine similarity between the tokens of the reference and the generated summary. We use the BERT-base model for evaluating the BERTScore.
    \item \textbf{FEQA} \cite{durmus-etal-2020-feqa} is a question-answer based metric. The pretrained QA model generates question-answer pairs from the summary, and later a model tries to find answers from the source document. The number of correct answers found assesses the faithfulness of the system.
    \item \textbf{Pyramid Score} \cite{pyramid-m} uses Semantic Content Unit (SCUs) to find correlation between the generated summary and the optimal summary. We use automated variant to compute Pyramid score.
\end{itemize}

\subsection{Experimentation Environment}
All experiments performed in this study use a 2 X 48GB A6000 GPU and a 32 core AMD CPU coupled with 128GBs of RAM. The GPU uses CUDA 10/11 driver environment with cuDNN-7 installed. We use the standard Pytorch 1.8 package in our main environment. For baselines, We preserve all the hyperparameters provided in the original works. Our experiments also share similar baseline hyperparameters.

\begin{figure*}

    \centering
    \scalebox{0.4}{
    \includegraphics[trim={1cm 0 0 0},clip]{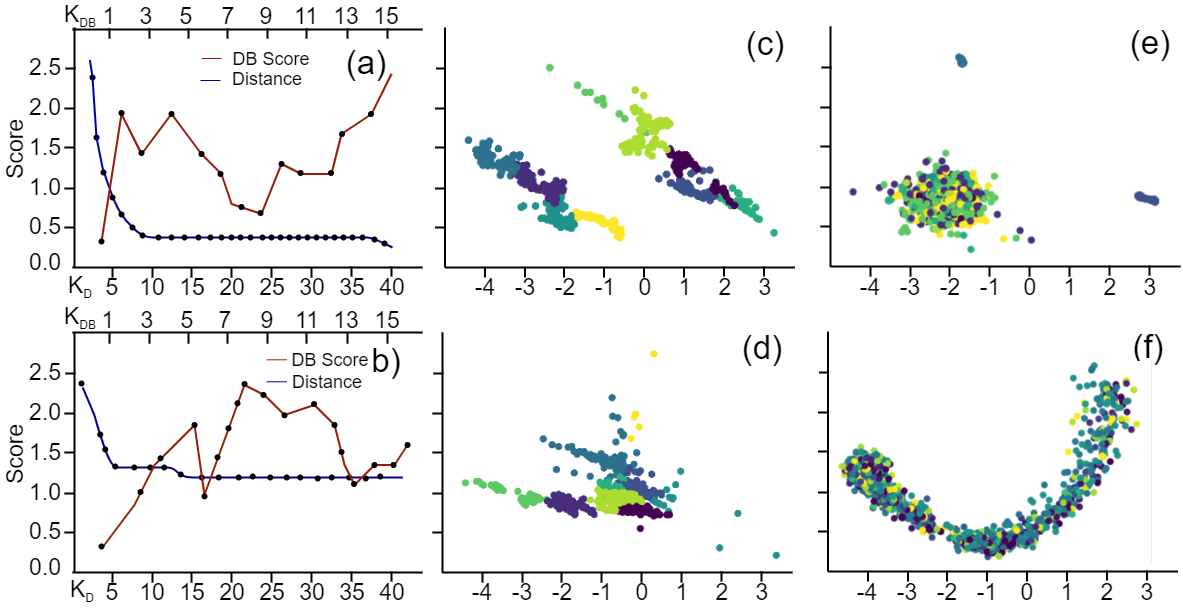}}
\caption{Visualizations of (a) Elbow and Davies Bouldin score for LSTM models, (b) Elbow and Davies Bouldin score for Transformer models, (c) LSTM embedding vector visualization using PCA, (d) LSTM encoder vector visualization using PCA, (e) Transformer embedding vector visualization using PCA, (f) Transformer encoder vector visualization using PCA.}
\label{fig:cluster_map}
\end{figure*}


\section{Bias in Summarization systems} \label{sec:biasstudy}

In this section, we discuss how the representation bias is characterized, quantified and exploited for quantitative and qualitative performance.

Text summarization is challenging due to the lengthy source document and varying thematic topics across a single-source document. The target summary also has to contribute equally in terms of capturing the critical information to make deep learning systems learn to extract and map the critical sections of information. Compared to other tasks such as machine translation, dialogue generation, and next token prediction, where the systems tend to generate output based on a local context window and a central theme, summarization systems need to broaden the context to a global level both within and across documents. To learn the diverse information, summarization systems should portray the spread of information in the vector space to show that deep learning systems can learn various representations and are not focused on local context windows or omitting the theme of the input document.

\subsection{What is Representation Bias?}
A set of samples in a dataset represents the variation of the population. Any form of data which lacks the consideration of outliers, uneven data points, diversity and anomaly factors give rise to representation bias. For example, the paucity of varying geographical presence in the Imagenet \cite{deng2009imagenet} dataset formulates a representation bias towards the white population. Similarly, in case of abstractive summarization, the scarcity of data points portraying as outliers, uneven themes and non-templatic target summaries lead to representation bias towards a specific set of data samples leading to degraded performance. This also abstains deep learning based systems from generalizing across all genres of datasets. We study this representation bias for the task of abstractive summarization and exploit it to gain improvements with qualitative data points.  

\subsection{Characterizing and Modeling Bias in Training Dynamics} \label{sec:biasstudy}
Inspired by the study of biases \cite{10.1145/3404835.3462846,zhong-etal-2019-closer,dey-etal-2020-corpora}  portraying that the current deep learning based summarization systems show qualitative issues like repetitions, unfaithfulness and non-coherence, we dive deep into the reasoning behind the same. 

Given a source document and the corresponding target summary, we train an end-to-end deep abstractive summarization system keeping the batch size as $1$. The corresponding embeddings and the encoder outputs learned at each time step are saved. We save weights at every epoch to have the representation of the trained embeddings and the encoder representation learned. After the convergence of the system, we cluster the embedding representations and the encoder weights. The intuition behind clustering is to understand the distance or the diversity of the data points in the vector space. The idea is that the data points should be spread far apart to represent the diversity of varied examples in the training sample.

For clustering of embedding and encoder weights, we use the KNN+ algorithm. We consider the Elbow \cite{8549751} method and the Davies Bouldin \cite{4766909} method to find the optimum number of the clusters for both the embedding and the encoder weights. For the LSTM-based systems, the optimal number of clusters for the embeddings by elbow method as well as the Davies Bouldin method came out to be $9$, as shown in Fig. \ref{fig:cluster_map}(a). We visualize the formulated clusters for embeddings by reducing the dimensions using Principal Component Analysis (PCA)  in Fig. \ref{fig:cluster_map}(c). 

 \begin{figure}
    \centering
    \scalebox{0.53}{
    \includegraphics[trim={8.6cm 4.5cm 5cm 3cm},clip]{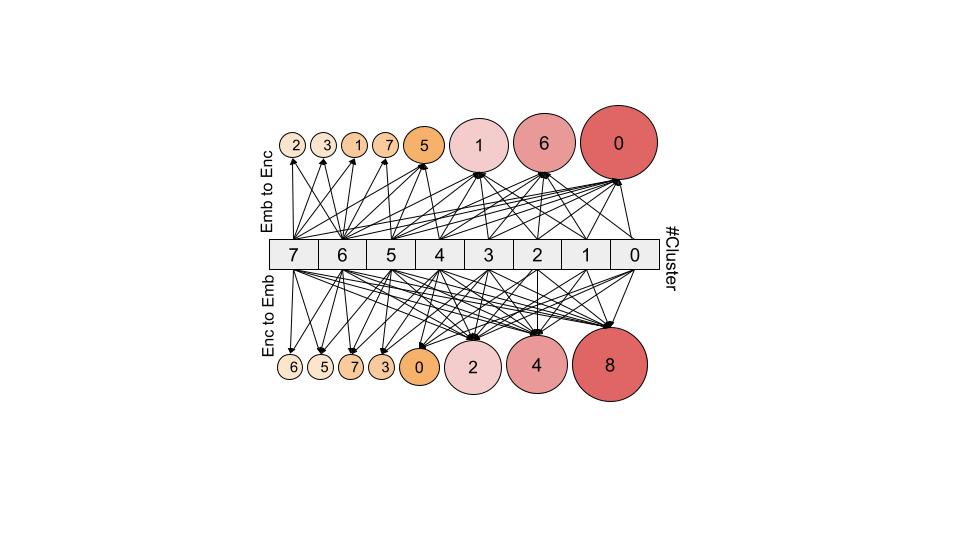}}
\caption{Visualizations showing the mapping of clusters for LSTM-based systems from embedding-to-encoder (array to upper cluster cloud) and encoder-to-embedding (array to lower cluster cloud). }
\label{fig:pgclusters}
\end{figure}

We also follow the same steps for the encoder weights. The optimal number of clusters for encoder weights is also $9$ using both the Elbow and the Davies Bouldin methods. Fig. \ref{fig:cluster_map}(d) shows the encoder representation by reducing the dimension size using PCA. The visualization of encoder weights shows the reduced data points when the number of clusters is set to $9$. From the visualizations, we infer that the vector space representation of both the embedding and the encoder outputs is not diverse in the space and is heavily saturated. Comparatively, encoder weights show a more diverse spread on the embedding space than the embeddings; yet the data points look heavily saturated for both the representations.

Similarly, for the Transformer-based systems, we again use the KNN+ algorithm to perform clustering of embeddings and the encoder weights. To find the most optimal number of clusters, we use the Elbow  and the Davies Bouldin methods. From both the methods, the most optimum number of clusters came out to be $5$. Keeping $5$ as the cluster parameter, we visualize the embedding data points. However, the visualizations obtained came out to be heavily clustered and saturated at a single point. The points that are at the farthest position are three times the average Euclidean distance, making them outliers. We take the second best optimal cluster count as $14$ and visualize the corresponding embedding points. 

We visualize the corresponding data points using PCA in Figs. \ref{fig:cluster_map}(e) and \ref{fig:cluster_map}(f), inferring that the Transformer-based systems show a more diverse representation of data points at the encoder side than any LSTM-based systems. 
\begin{figure}
    \centering
\scalebox{0.48}{
    \includegraphics[trim={10.5cm 4.3cm 5cm 3cm},clip]{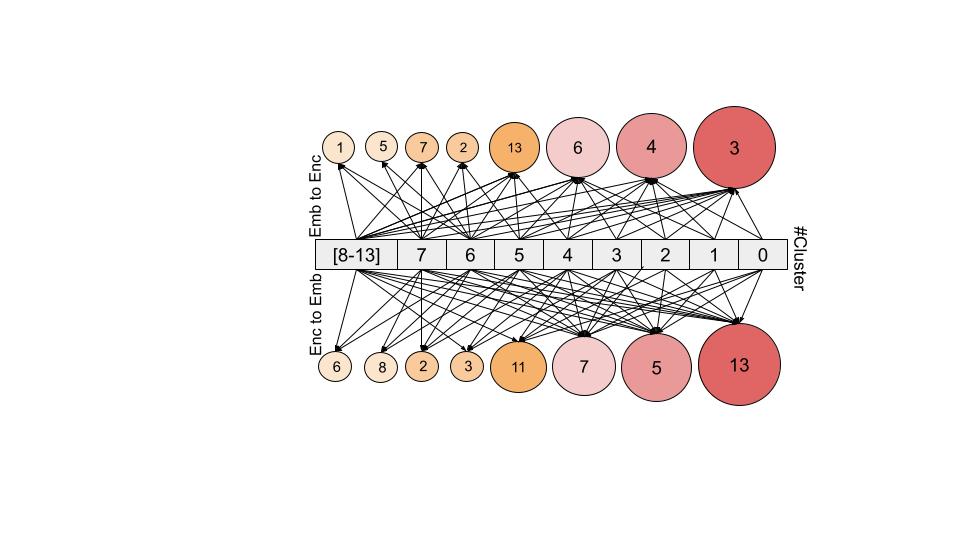}}
\caption{Visualizations showing the mapping of clusters for Transformer-based systems from embedding-to-encoder (array to upper cluster cloud) and encoder-to-Embedding (array to lower cluster cloud).}
\label{fig:transclusters}
\end{figure}

To better understand the delineation between the embeddings and the encoder representation, we map the data points originating from the embedding clusters and falling in the encoder clusters for the LSTM-based models shown in Fig. \ref{fig:pgclusters}. For embedding-to-encoder map, we see that most of the data points fall into three major encoder clusters, namely $\#0$, $\#6$ and $\#1$. For cluster $\#0$ in the encoder, data points map to clusters $\#0$ at $71\%$, and cluster $\#7$ at $21\%$, while the rest of the embedding majority is the same. For the encoder-to-embedding mapping, most data points from the encoder cluster $\#3$ fall in embedding clusters $\#4$, $\#2$, and $\#8$ except for the encoder cluster $\#0$, where $\#2$ is the majority cluster. Similarly, for the Transformer-based systems, a similar pattern is observed (Fig. \ref{fig:transclusters}) in the embedding-to-encoder and the encoder-to-embedding mappings. For embedding-to-encoder, the majority of the data points fall in clusters $\#3$, $\#4$, and $\#6$ with minor variations. Similarly, for the encoder-to-embedding mapping, the majority falls to clusters $\#13$, $\#7$, and $\#5$ with minor variability across classes. This taxonomy of cluster data point mapping between the embedding weights and the encoder weights shows that the deep learning based systems learn a similar representation irrespective of the training style or the architecture. This saturation of weights forming limited clusters thwarts the performance and makes the model learn a fixed-type of output template that is repeated in the generated summary irrespective of the input document.

\begin{table*}[!htp]\centering
\small
\scalebox{1}{
\begin{tabular}{l|c|cccccccc}\toprule
Dataset &No. of samples &Rouge-1  &Rouge-2 &Rouge-L F1 &Rouge-L &BERTScore &FEQA &Pyramid \\\midrule
\multirow{5}{*}{CNN/Dailymail} &Random 10K &34.57 &14.83 &31.34 &30.29 &22.63 &12.79 &0.28 \\\cmidrule{2-9}
&10K &39.49 &17.84 &36.38 &36.31 &26.47 &16.52 &0.34 \\\cmidrule{2-9}
&40K &39.96 &17.66 &36.58 &36.59 &26.44 &16.48 &0.34 \\\cmidrule{2-9}
&100K &40.18 &17.74 &36.67 &36.73 &26.49 &16.49 &0.35 \\\cmidrule{2-9}
&Complete &40.13 &17.76 &36.84 &36.79 &26.57 &16.57 &0.36 \\\midrule
\multirow{5}{*}{Multinews} &Random 10K &34.1 &10.24 &13.64 &13.4 &21.54 &9.33 &0.23 \\\cmidrule{2-9}
&10K &42.76 &13.86 &17.07 &17.02 &33.88 &13.24 &0.27 \\\cmidrule{2-9}
&20K &42.89 &13.94 &17.09 &16.95 &33.89 &13.24 &0.26 \\\cmidrule{2-9}
&40K &43.41 &13.91 &17.13 &17.08 &33.83 &13.25 &0.27 \\\cmidrule{2-9}
&Complete &43.49 &14.02 &17.21 &17.16 &34.21 &13.27 &0.28 \\\midrule
\multirow{5}{*}{CQASumm} &Random 10K &22.41 &3.12 &14.62 &14.52 &18.41 &7.02 &0.13 \\\cmidrule{2-9}
&10K &29.67 &4.87 &19.64 &19.44 &22.12 &9.64 &0.21 \\\cmidrule{2-9}
&30K &29.41 &4.86 &19.77 &19.67 &22.04 &9.65 &0.19 \\\cmidrule{2-9}
&50K &29.38 &4.91 &20.17 &19.92 &22.06 &9.68 &0.22 \\\cmidrule{2-9}
&Complete &31.47 &5.02 &20.28 &20.24 &22.16 &9.71 &0.24 \\\midrule
\bottomrule
\end{tabular}}
\caption{Benchmarking scores over various subsampled data on the metrics -- Rouge-1, Rouge-2, Rouge-L, Rouge-L F1, BERTScore, FEQA, and Pyramid score for the datasets -- CNN/Dailymail, Multinews and CQASumm.}\label{tab:performance}

\end{table*}
\subsection{Exploiting Bias for Performance} \label{sec:exploit}
As the number of clusters and the representations are dense and saturated for both the embedding and encoder weights, we sample the data points from each cluster to obtain diverse representations to make the systems more generalizable, easier to train, and perform better across the dataset. The subsampling helps in the data distillation as to use only the essential data points during training.

\begin{figure}[!h]
    \centering
    \scalebox{0.63}{
    \includegraphics{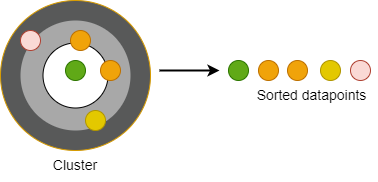}}
    \caption{A cluster showing the samples with respect to the distance from the center. The datapoints are rearranged according to the distance from minimum to maximum.}
    \label{fig:subcluster}
\end{figure}

For subsampling of the data points, we extend the study of \cite{9578663} to filter the data points. For every data point in the training set, we use the cosine similarity based KNN+ algorithm. The cluster $C_i$ is formulated using
\begin{equation}
    C_i = \kappa (f_i,F,k)
\end{equation}
 where $F = feat(X)$ is the feature matrix, and $k$ is the nearest neighbour. Now, for the cluster $C_i$, we define another sub-cluster $S_i$,
 \begin{equation}
     S_i = \gamma(C_i)
 \end{equation}
with $f_i$ as the centre of cluster $C_i$  and $f_j$ forming as the neighbor of $f_i$. The similarity of all the points is computed with respect to the centre as a vector $\{S_{i,j}\}^k_{j=1}$. The data points in $S_i$ are sorted concerning the similarity function as shown in Fig. \ref{fig:subcluster}. The subsamples based on the upper limit are extracted uniformly from each $S_i$.

The training data points obtained from the sorted clustered samples are benchmarked by finetuning them over the BART \cite{lewis-etal-2020-bart} language model. We finetune BART over the complete CNN/Dailymail, Multinews and CQASumm dataset, and obtain the overall benchmark results. We then perform the subsampling to obtain $10K$ subsamples for the CNN/Dailymail data, Multinews and CQASumm. We finetune BART on the $10K$ subsampled data and compare the performances with the complete data (Table \ref{tab:performance}). 

We observe that for CNN/Dailymail data, the $10K$ subsampled instances attain a $39.49$ R1 and $36.38$ RL, an improvement of $+4.92$ R1 and $+5.04$ RL over the Random $10K$ and only $0.64$ R1 and $0.46$ RL lower than the performance on the complete dataset. Similarly, FEQA and the Pyramid score gain an improvement of $+3.73$ and $+0.06$, respectively against the random $10K$ samples showing that the qualitative performance can be achieved with just $10\%$ of data without compromising on the quantitative performance. Similarly, when $40k$ subsampled data is finetuned, the performance gain is only $0.17$ R1 and $0.26$ RL lower than the complete dataset. For Multinews, the subsampled $10K$ data produces $42.76$ R1 and $43.49$ RL, a reduction of just $0.73$ R1 and $0.14$ RL while giving an improvement of $+0.4$ FEQA and $+0.01$ Pyramid score. We notice similar trend for the CQASumm dataset on which the gain is $+0.04$ FEQA and $+0.02$ Pyramid score (Table \ref{tab:performance}).

\section{Analysis}

In the previous section, we assess the diversity of the embedding weights and the encoder weights in the vector space learned during the model training. Our intuition is that the deep learning based models should learn a diverse representation so as to handle variable source input. We analyze these characteristics for both  types of the summarization systems -- LSTM-based and Transformer-based. Our study indicates that both types of models often learn similar weight matrices across the embedding as well as in the encoder vector space shunning the variable input representation. Even though the embeddings can represent the variation in the vector space but the summarization system, during its training, maps these data points to a saturated point at the encoder side. From Fig. \ref{fig:pgclusters}(c), we observe that the embeddings show large variability; however, the encoder space shows the points getting saturated. In case of Transformers, the embedding weights are saturated; however, Fig. \ref{fig:transclusters}(f) shows that the embedding points are again mapped to a saturated continuous sequence of clusters. This indicates that the current systems, irrespective of the deviation in the input source, try to learn similar weights at the encoder side, giving a templatic and repetitive output in the generated summaries.
After the analysis, we filter out the data points using the hierarchical clustering algorithm discussed in the previous Section and finetune them over the BART \cite{lewis-etal-2020-bart} language model. When compared with the performance of the systems trained on the complete data, the subsampled $10K$ finetuned systems stand comparable while the performance is at par against those finetuned on the randomly-sampled data. This shows that an uncontrolled urge to extract more data for the models is not the right way; limited qualitatively labelled data can provide the model with sufficient information to make it learn diverse representations, generalize across datasets and easier to train.
\section{Discussion}
We discuss some of the insights obtained from the analysis performed over the biases characterized.

\textbf{Does deep learning model favour generalization? }
Deep learning models, in general, exploit the biases to learn the representations between the source data and the target data. The relations obtained act as weights to map the contextual relationship for providing inference on the unseen data. However, the exploitation does not always lead to better benchmark scores. For a given problem, the datasets are often collected from a crowdsourced platform that may affect the quality of the dataset due to various reasons such as absence of moderation, limited themes, and more engagement over the trending topics. These limitations make the data less robust and limited to be usable over a particular genre or a problem. Often the proposed models are benchmarked on limited datasets as they fail to perpetuate the performance across similar problems. 

\textbf{Does diversity in training samples help in generalizing the systems? } Deep learning based systems require a massive amount of data for training. The available training data is an example of the representation that the system should accept during inference for generating output. The system learns the deviations in the individual train instance to adjust its parameters in order to handle variability and diversity during inference. The diversity in the training data helps the system learn these variations. However, the degree of these variations poses a question of whether to treat the sample as a variation or an outlier. If treated as a variation, the outlier sample may inflate the error metric, causing the system diverge from its intended goal. The threshold of what a system should treat as an outlier and a regular sample acts as a foundation of diversity in the training dataset. Since most of the training examples lie very close to each other due to similar topics or themes, the threshold of a system to consider a data instance as an outlier decreases, making the model treat the diverse training examples as outliers and learn a limited representation as to the actual mapping.

\textbf{Can the biases be exploited for performance?}
Biases in summarization systems have been exploited to favour performance. \cite{10.1145/3404835.3462846} exploited the lead bias in the news datasets to attain high quantitative benchmarks. News datasets contain a majority of the vital information in the first few paragraphs, and the systems can attain this really well, leading to high performance. \cite{fabbri-etal-2021-improving} used data upsampling and transfer of weights from a language model to make models perform with a few shot transfer learning. For a dataset, each model exploits characteristics which may or may not be known to the researchers to visualize and tweak. Our work focuses heavily on the aspect that the biases obtained during the training phase can be exploited to attain similar performance as obtained when a model is trained on the complete dataset.

\textbf{Can the biases be rectified for datasets?}
What actually constitutes a bias? The definition of bias is subjective and can not be 
defined as a function of either a dataset or a model. For some, a bias can be towards gender-imbalanced inferences, and for others, a bias can be a generation of templatic output for different source inputs. Each type of bias requires a specific solution and a prescription either in the data or the model. As discussed by \cite{fabbri-etal-2021-improving}, a single system can not rectify all errors. Hence, a study is needed to find the forms of biases that either help in terms of required improvements or limit the performance of the model qualitatively or quantitatively. In addition, biases can also be a result of internal factors like data shuffling, overfitting or external factors like a biased opinion of a crowdsourced platform. However, in this study, we mainly focus on the biases that thwart the performance of the model qualitatively. We study the system's training dynamics and the representations learned by the system at the embedding and encoder levels. Later, we exploit these biases to gain improvements over the current summarization datasets in terms of faithfulness and qualitative inference.

\section{Conclusion}
In this paper, we characterized the representation bias propagating from the dataset into the model. We discussed the reasoning which thwarts the performance of the deep learning based models on varied datasets. We also discussed how the representation of the data in the embeddings and the representation of the model in the encoder space are heavily skewed and do not show diversity or variability. We used hierarchical clustering to filter out data points that show similar statistical features as a complete dataset and showed that the subsampled data can achieve similar performances when benchmarked against the complete data. We exploited the representation bias to gain performance on a subsampled dataset over various metrics. We discussed why the current datasets and summarization systems fail to generalize across data. 

\bibliographystyle{IEEEtran}
\bibliography{bibliography}

\vfill

\end{document}